%% file: main.tex
\documentclass[journal]{IEEEtran}
\usepackage{preambles}

\begin{document}
\maketitle
\input{abstract}
\input{1-introduction}
\input{2-methods}
\input{3-results}
\input{4-discussion}
\input{5-conclusion}

\bibliography{refs}

\end{document}

%% file: abstract.tex
\begin{abstract}
\textit{Goal:} The countermovement jump (CMJ) is commonly used to measure lower-body explosive power. This study evaluates how accurately markerless motion capture (MMC) with a single smartphone can measure bilateral and unilateral CMJ jump height. \textit{Methods:} First, three repetitions each of bilateral and unilateral CMJ were performed by sixteen healthy adults (mean age: 30.87$\pm$7.24 years; mean BMI: 23.14$\pm$2.55 $kg/m^2$) on force plates and simultaneously captured using optical motion capture (OMC) and one smartphone camera. Next, MMC was performed on the smartphone videos using OpenPose. Then, we evaluated MMC in quantifying jump height using the force plate and OMC as ground truths. \textit{Results:} MMC quantifies jump heights with ICC between  0.84 and 0.99 without manual segmentation and camera calibration. \textit{Conclusions:} Our results suggest that using a single smartphone for markerless motion capture is promising.
\end{abstract}


\renewcommand\IEEEkeywordsname{Impact Statement}
\begin{IEEEkeywords}
    Countermovement jump height can be accurately quantified using markerless motion capture with a single smartphone, with a simple setup that requires neither camera calibration nor manual segmentation.
\end{IEEEkeywords}

%% file: 1-introduction.tex
\section{Introduction}
\label{sec:introduction}
The countermovement jump (CMJ) is commonly used to measure lower-body explosive power and is characterised by an initial downward movement of the centre of mass (COM), known as \emph{countermovement}, before toe-off \cite{cmj_review}. Performance assessment with CMJ often involves motion capture and measurement of metrics such as peak velocity and vertical jump height. Traditionally, motion capture is performed using wearable sensors, optical motion capture (OMC) equipment and force plates, which are highly accurate. However, compared to smartphones, they are relatively expensive, not readily portable, and their operation requires some level of technical instruction. In addition, OMC requires physical body markers, which can be affected by skin and clothing artefacts. Moreover, wearable sensors, physical markers, and the awareness of being under observation may alter the real performance of subjects  \cite{wade_needham_mcguigan_bilzon_2022, geh_beauchamp_crocker_carpenter_2011}.

\begin{figure*}[ht]
    \centering
    \includegraphics[width=1\textwidth]{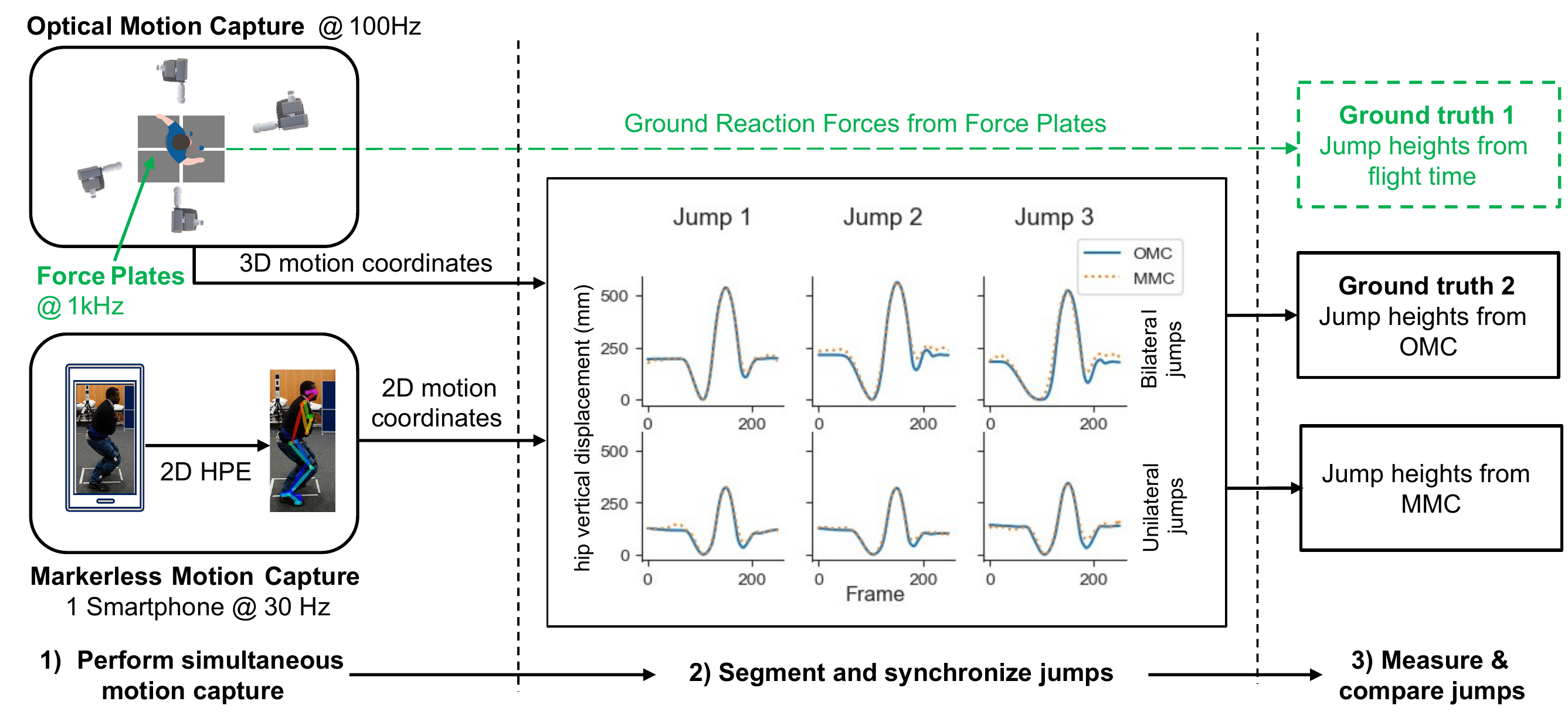}
    \caption{Experiment setup showing simultaneous motion capture, preprocessing, and comparison with ground truths.}
    \label{fig:experiment_setup}
\end{figure*}

Recent advances in computer vision research have enabled \textit{markerless motion capture} (MMC) from videos. MMC often relies on human pose estimation (HPE) algorithms such as AlphaPose \cite{fang2017rmpe}, OpenPose \cite{openpose}, and DeepLabCut \cite{mathis2018deeplabcut}. These MMC techniques have shown potential to replace OMC, especially since smartphones are ubiquitous. However, there is still a lot to be done in evaluating the accuracy and usability of MMC.

Existing MMC approaches can be categorised based on \textit{capture plane} (2D or 3D) and \textit{number of cameras} (multi- or single-camera). 2D monocular (single-camera) techniques have been used for quantifying limb kinematics during underwater running \cite{under_water_running} and sagittal plane kinematics during vertical jumps \cite{outside_the_lab}. However, these works rely on deep learning approaches, where the generalisation ability depends on the size and diversity of the data and the model architecture. For example, trained athletes, casual trainers, and rehabilitation patients will exhibit different performance ranges. Since collecting large quantities of representative data is difficult, we take an alternative approach here, a quantitative approach, and we focus on the ease of deployment in practice and ease of use. The \emph{MyJump2} \cite{my_jump} app has been deployed for measuring jump height using a single smartphone. However, it requires manual selection of jump start and end frames. Previous researchers have performed 3D MMC using multiple cameras \cite{nakano2020evaluation, corazza_mmc}. However, the 3D multi-camera approach requires careful calibration and reconstruction of 3D poses from multiple 2D camera angles, which is not feasible for wide deployment in practice.

Therefore, we evaluated a single-smartphone-based MMC in measuring bilateral and unilateral countermovement jump height. \textbf{Our main contributions are:}
\begin{enumerate}
    \item We use a simple setup with a single smartphone, with no strict requirements on view perpendicularity and subject's distance from the camera. This is a more realistic application setting where MMC is used outside the lab, without specialised equipment.
    \item We show how to exploit gravity as reference for pixel-to-metric conversion as proposed in \cite{gravity_ref}, removing the need for reference objects or manual calibration.
    \item We analyse how accurately MMC measures jump heights compared with OMC and force plates.
    \item We discuss situations in which MMC could be potentially useful.
\end{enumerate}

%% file: 2-methods.tex
\section{Materials and Methods}
\subsection {Participants}
Sixteen healthy adults (mean age: 30.87$\pm$7.24 years; mean BMI: 23.14$\pm$2.55 $kg/m^2$) volunteered to participate in this study. The dominant foot of each participant was determined based on the foot with which they kick a ball \cite{kick_a_ball}. Each participant signed the informed consent form approved by the Human Research Ethics Committee of University College Dublin with Research Ethics Reference Number LS-C-22-117-Younesian-Caulfield.

\subsection {Tasks}
After a five-minute warm-up, each participant performed three repetitions each of CMJ bilateral (BL) and unilateral (UL) while simultaneous motion capture was performed using force plates, OMC, and MMC (Fig. \ref{fig:experiment_setup}).

\subsection {Apparatus}
\subsubsection {Force Plate}
\label{sec:force_plate}
\begin{figure}
    \centering
    \includegraphics[width=1\linewidth]{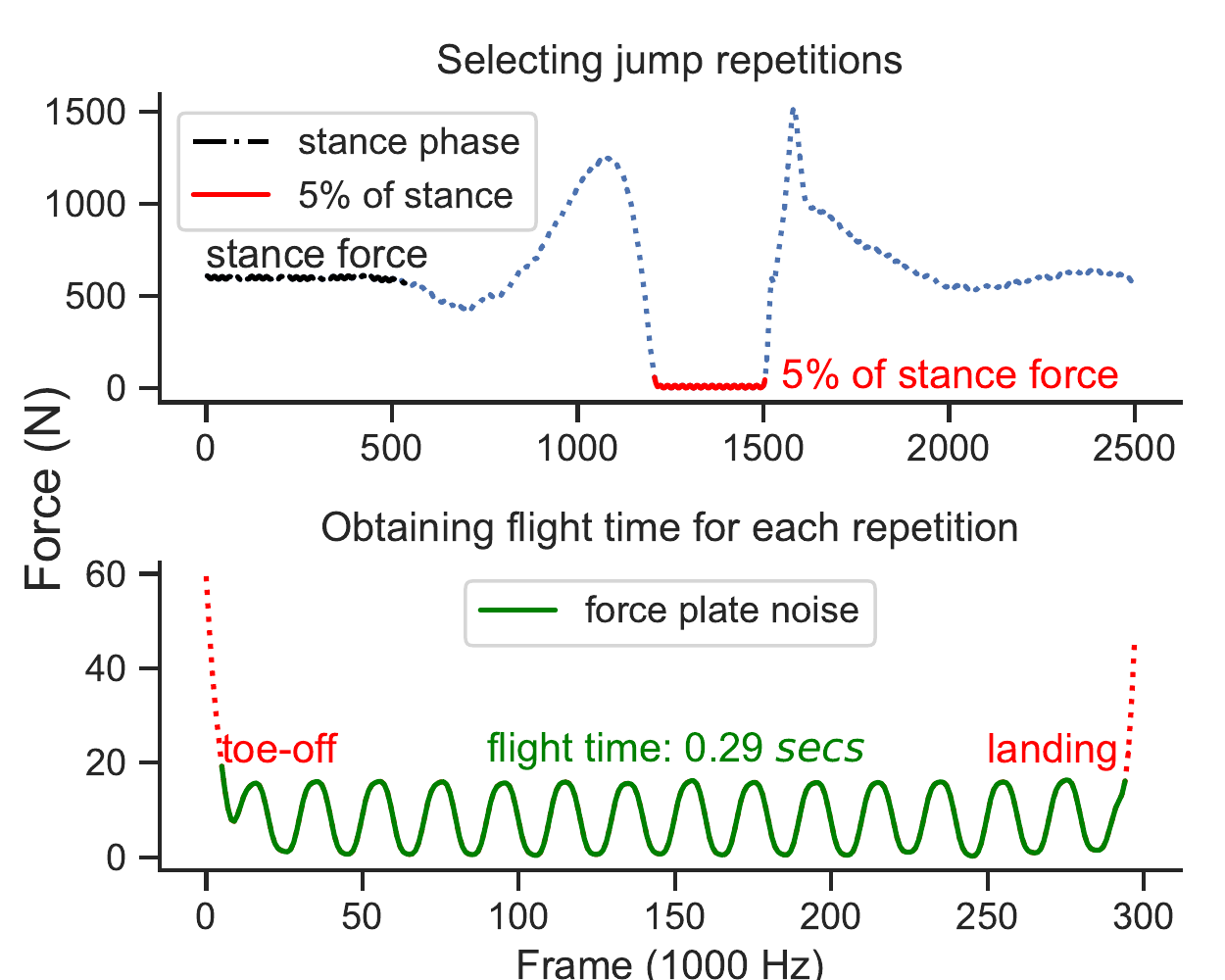}
    \caption{Obtaining flight time from the force plate, shown for a single repetition.}
    \label{fig:fp-ft}
\end{figure}

AMTI\footnote{Advanced Mechanical Technology, Inc (\url{https://amti.biz})} force plates sampling at 1000 Hz were used as the first ground truth. To obtain the flight time $T_f$ for each jump, we first identified jump repetitions by selecting force values less than 5\% of the force in the stance phase. Then, for each selected repetition, we identified the toe-off and landing forces as sudden force changes relative to the noise of the unloaded force plate, thereby obtaining $T_f$ more precisely (Fig.~\ref{fig:fp-ft}). We obtained the jump height in centimetres as
\begin{equation}
    h = 100gT_f^2/8
\end{equation}
where $g$ is the acceleration due to gravity \cite{gavin}.

\subsubsection {Optical Motion Capture}
\label{sec:omc}
Optical motion capture was performed using four synchronised CODA\footnote{Charnwood Dynamics, UK (\url{https://codamotion.com})} 3D cameras sampling at 100 Hz and synchronised with the force plate. Four clusters, each consisting of four light-emitting diode (LED) markers, were placed on the left and right lateral sides of the thigh and shank (Fig.~\ref{fig:noisy}). Moreover, six LED markers were placed on the anterior superior iliac crest (anterior and posterior), and greater trochanter (left and right). Three LED markers were attached to the lateral side of the calcaneus and on the first and fifth metatarsals of the dominant foot.

For a motor task with duration $T$ seconds and $K$ tracked joints, CODA outputs a sequence of 3D coordinates \(\{(x_i^t,y_i^t,z_i^t ) | i=1,...,K;t=1,...,100T\}\) in \textit{millimetres}; where $z$ is the vertical axis, and 100 is the sampling rate.

\subsubsection {Markerless Motion Capture}
\label{sec:mmc}
Markerless motion capture was performed in the side view using one Motorola G4 smartphone camera with a resolution of 720p and a frame rate of 30 frames per second (fps). The smartphone was placed on a tripod perpendicular to the dominant foot of the participant. We placed no strict requirements on camera view perpendicularity and distance to the participant. However, we ensured that the camera remained stationary and participants remained fully visible in the camera view.

To obtain motion data from the recorded videos, we performed 2D HPE using OpenPose \cite{openpose}. The HPE algorithm outputs a sequence \(\{(x_i^t,y_i^t,c_i^t )| i=1,...,K;t=1,...,30T\}\), where 30 is the frame rate, \((x_i^t,y_i^t)\) are the 2D coordinates in \textit{pixels}, and \(c_i^t \in [0,1]\) is the probability for joint $i$ in frame $t$.

\subsection{Data Preprocessing}
\label{sec:data-preprocessing}
During preprocessing, we performed denoising, segmentation, resampling, and rescaling.

\begin{figure}[ht]
    \centering
    $\begin{array}{cc}
         \subfloat[]{%
          \includegraphics[width=1\linewidth]{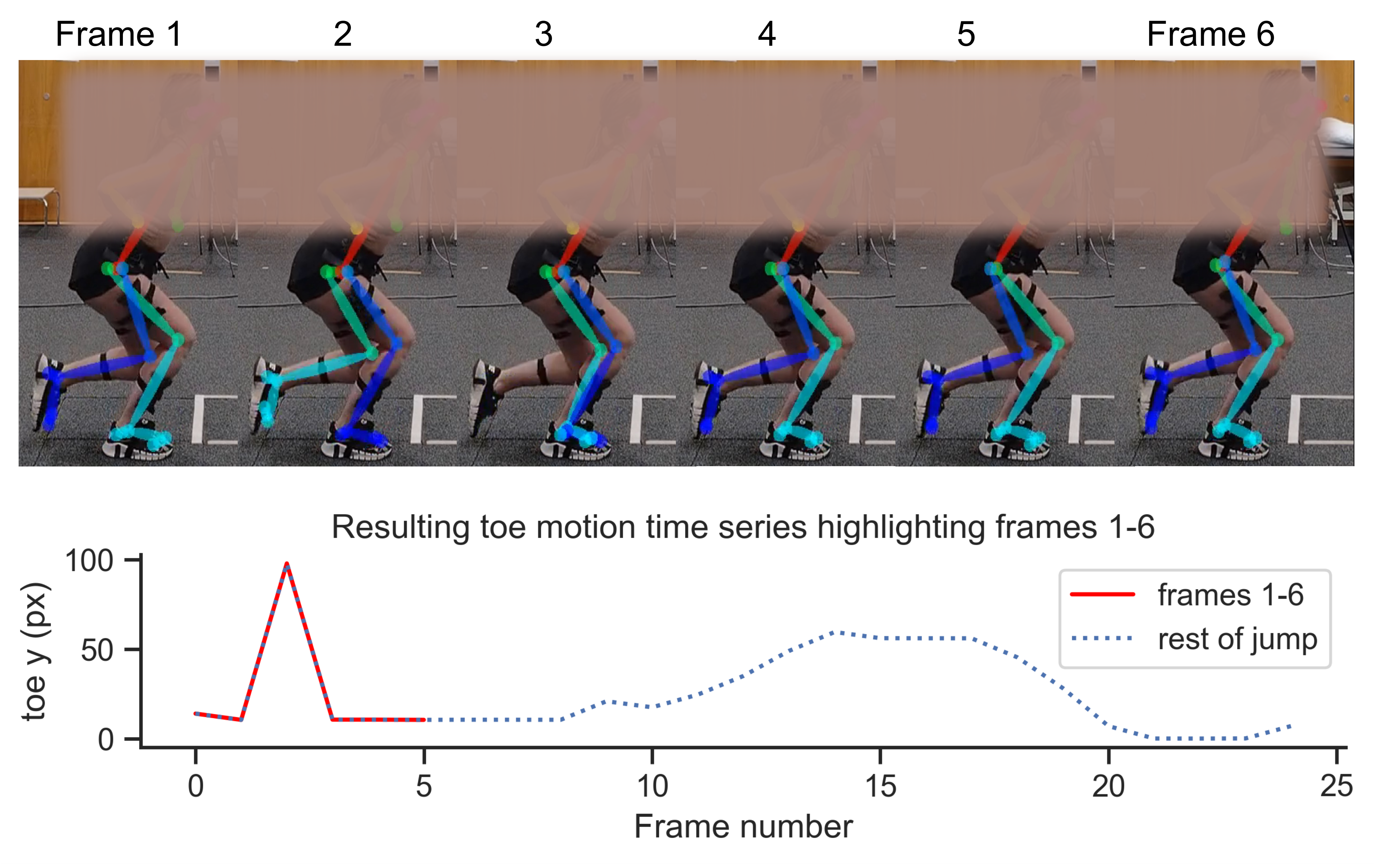}
        } \\
        \subfloat[]{%
          \includegraphics[width=1\linewidth]{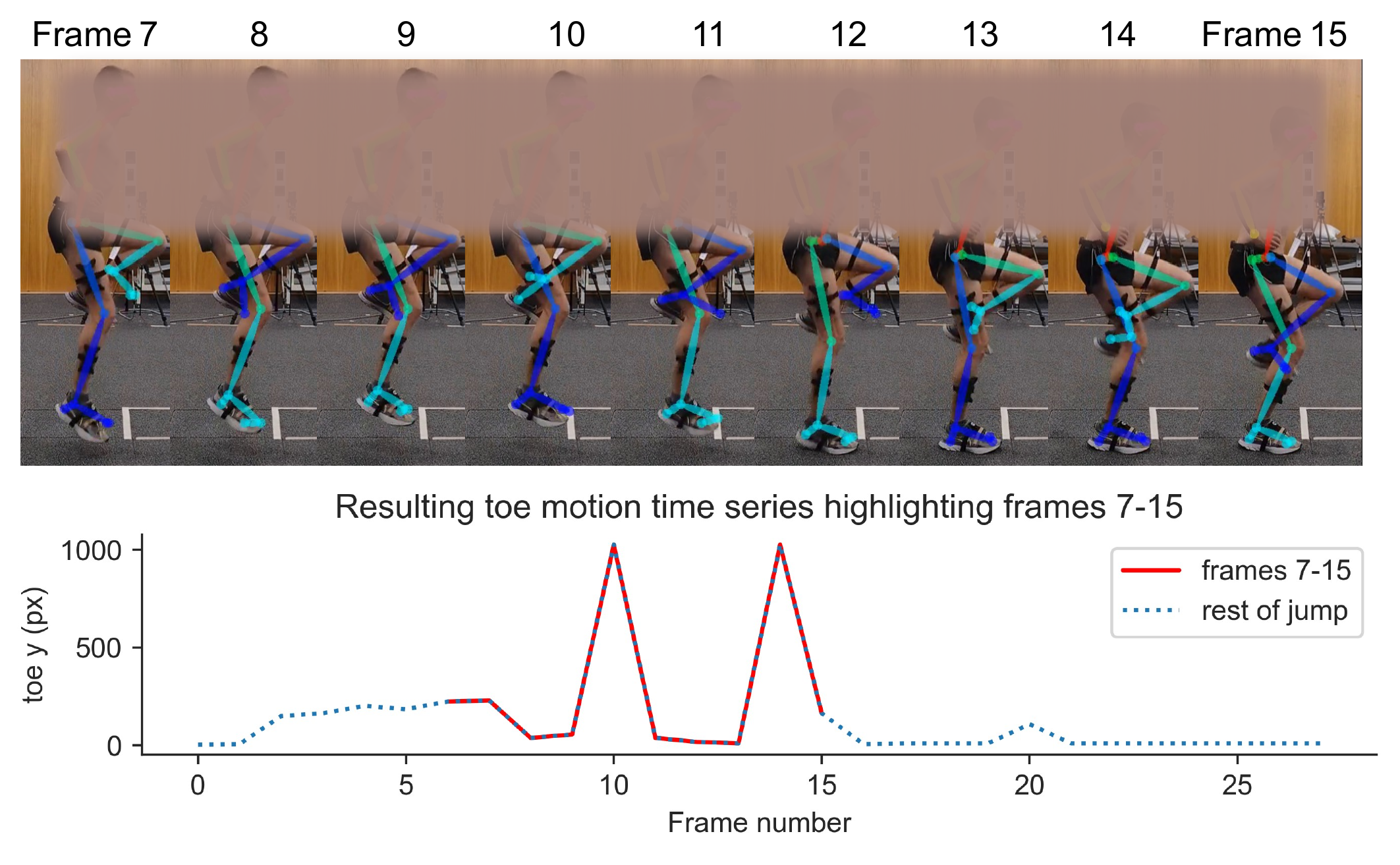}
        }
    \end{array}$
    \caption{Examples of noise in unilateral jumps during pose estimation as seen by observing the limb heatmap colors. (a) In frame 2, the left and right limbs are swapped. In frame 3, the right limb is wrongly detected as two limbs. (b) A failure case showing movements that are not characteristic of countermovement jumps.}
    \label{fig:noisy}
\end{figure}

\subsubsection{Denoising}
As shown in Fig.~\ref{fig:noisy}(a), occasional false detections in pose estimation appear as spikes on the motion time series. In most cases, these spikes could be removed by smoothing. However, 19 unilateral jumps such as Fig.~\ref{fig:noisy}(b) showed uncharacteristic movements and were removed as failure cases. To avoid filtering out important motion data, we performed smoothing of the OMC and MMC time series using z-score smoothing \cite{aderinola}, proposed specifically for spike removal in motion sequences, and a second-order Savitzy-Golay \cite{savitzky1964smoothing} (Savgol) filter. The Savgol filter is known to smooth data with little distortion \cite{Guin2007MovingAA}, and we chose a window size of 21 to preserve the main maxima and minima of the time series for accurate segmentation.

\begin{figure}[ht]
    \centering
    \includegraphics[width=1\linewidth]{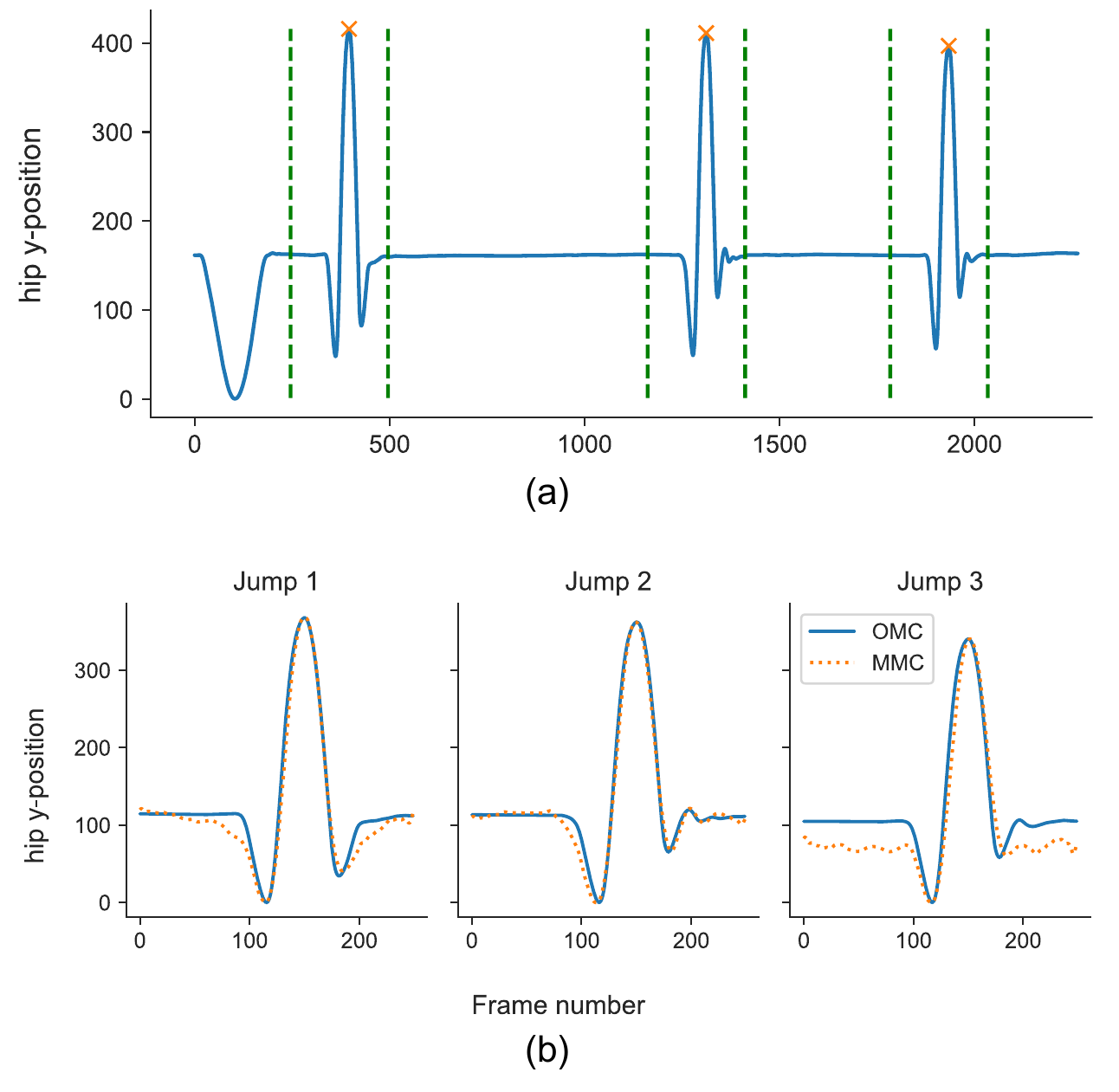}
    \caption{Segmentation of jumps repetitions. (a) Raw hip vertical motion signal with peaks and selected jump windows. (b) Segmented and synchronised jumps based on selected windows.}
    \label{fig:segment_and_sync}
\end{figure}

\subsubsection{Segmentation and Resampling}
Each jump repetition is characterised by a dominant peak corresponding to the maximum vertical height attained by the hip (Fig.~\ref{fig:segment_and_sync}). Using these peaks as reference, we segmented each jump with a window $t$ \emph{secs} to either side of each peak, where $t$ is based on exercise duration and capture frequency. This enabled the synchronisation of OMC and MMC based on start and stop times for each task. After segmentation, we upsampled the MMC time series to match the length of the OMC time series using Fast Fourier Transform resampling \cite{fourier_resample}, which minimised distortion.

\subsubsection{Rescaling}
\label{sec:rescaling}
Two approaches were taken to rescale MMC from pixels (\emph{px}) to a metric scale, namely \emph{reverse minmax} (RMM) and \emph{pixel-to-metric} (PTM).

\begin{table*}[ht]
\centering
\caption{Jump Heights From Force Plate, OMC, and MMC}
\label{tab:all_jumps}
\begin{threeparttable}
    \begin{tabular}{l|cccc|cccc}
    \hline
     & \multicolumn{4}{c|}{\textbf{Mean bilateral jumps} (cm)} & \multicolumn{4}{c}{\textbf{Mean unilateral jumps} (cm)} \\
    ID & FP & OMC & RMM & PTM & FP & OMC & RMM & PTM \\
    \hline
    P01 & 23.81 & 26.95 & 26.37 & 27.91 & 14.32 & 17.93 & 15.06 & 18.09 \\
    P02 & 11.81 & 12.58 & 10.96 & 12.43 & 8.64 & 10.68 & 8.60 & 9.60 \\
    P03 & 16.46 & 18.49 & 17.51 & 18.86 & 11.06 & 13.70 & 11.51 & 10.12 \\
    P04 & 18.19 & 21.54 & 20.52 & 21.54 & E & E & E & F \\
    P05 & 15.79 & 16.15 & 15.48 & 17.37 & 10.01 & 16.57 & 15.33 & 11.94 \\
    P06 & 15.88 & 17.64 & 16.76 & 19.35 & 8.22 & 10.68 & 10.24 & 11.48 \\
    P07 & 11.73 & 13.10 & 11.50 & 13.79 & 7.56 & 10.28 & 9.19 & 9.72 \\
    P08 & 13.70 & 15.63 & 12.80 & 12.99 & 5.88 & 7.73 & 5.41 & 5.65 \\
    P09 & 18.71 & 25.45 & 24.19 & 24.01 & E & E & E & F \\
    P10 & 18.50 & 20.24 & 19.49 & 20.62 & 12.10 & 16.39 & 13.81 & 14.10 \\
    P11 & 28.99 & 32.09 & 30.03 & 31.39 & E & E & E & F \\
    P12 & 15.15 & 20.99 & 17.98 & 18.10 & 6.72 & 12.42 & 9.47 & 8.99 \\
    P13 & 26.93 & 28.96 & 26.90 & 26.47 & 15.80 & 17.37 & 14.38 & 15.26 \\
    P14 & 33.96 & 37.99 & 36.68 & 35.99 & 13.43 & 16.01 & 14.35 & 15.67 \\
    P15 & 45.22 & 55.65 & 54.51 & 50.94 & 21.33 & 25.64 & 24.45 & 23.25 \\
    P16 & 26.22 & 26.50 & 24.82 & 21.12 & E & E & E & F \\
    \hline
    Mean & 21.32$\pm$8.80 & 24.37$\pm$10.56 & 22.91$\pm$10.64 & 23.31$\pm$9.52 & 11.54$\pm$4.19 & 14.58$\pm$4.61 & 12.62$\pm$4.66 & 12.82$\pm$4.55 \\
    \hline
    \end{tabular}
    \begin{tablenotes}
        {\item Each value is the mean across three repetitions. \textbf{F}: Failure cases (Fig.~\ref{fig:noisy}). \textbf{E}: The corresponding FP, OMC, and RMM unilateral jumps are excluded from analysis.}
    \end{tablenotes}
\end{threeparttable}
\end{table*}

\textbf{\textit{Reverse MinMax (RMM)}} involved using OMC as reference to rescale MMC into metric \emph{mm}. This was done by applying MinMax on both OMC and MMC, and then rescaling MMC into \emph{mm} using the scaling factor obtained from OMC. Let vectors $\mathbf{p_{mm}}$ and $\mathbf{q_{px}}$ represent the OMC (in mm) and MMC (in px) time series respectively. We obtained $\mathbf{q^*} = \textsc{minmax}(\mathbf{q})$, where $\mathbf{q^*_i \in [0,1]}$, as
\begin{equation}
    \mathbf{q^*} = \left\{ \frac{\mathbf{q_{px}}_i - \textsc{min}(\mathbf{q_{px}})}{\textsc{max}(\mathbf{q_{px}}) - \textsc{min}(\mathbf{q_{px}})} \right\}
    \label{eq:minmax}
\end{equation}
where $i=1, ..., N$, and $N$ is the length of $\textbf{q}$. We then obtained $\mathbf{q_{px}}$ in \textit{mm} scale as
\begin{equation}
\begin{split}
    \mathbf{q}_{mm} = \{\mathbf{q^*_i}[\textsc{max}(\mathbf{p_{mm}})-\textsc{min}(\mathbf{p_{mm}})]\\
    +\textsc{min}(\mathbf{p_{mm}}) | i=1,..., N\}
    \label{eq:rescale}
\end{split}
\end{equation}
Since RMM requires OMC as reference, it can be used for evaluation purposes only.

\begin{figure}[ht]
    \centering
    \includegraphics[width=0.95\linewidth]{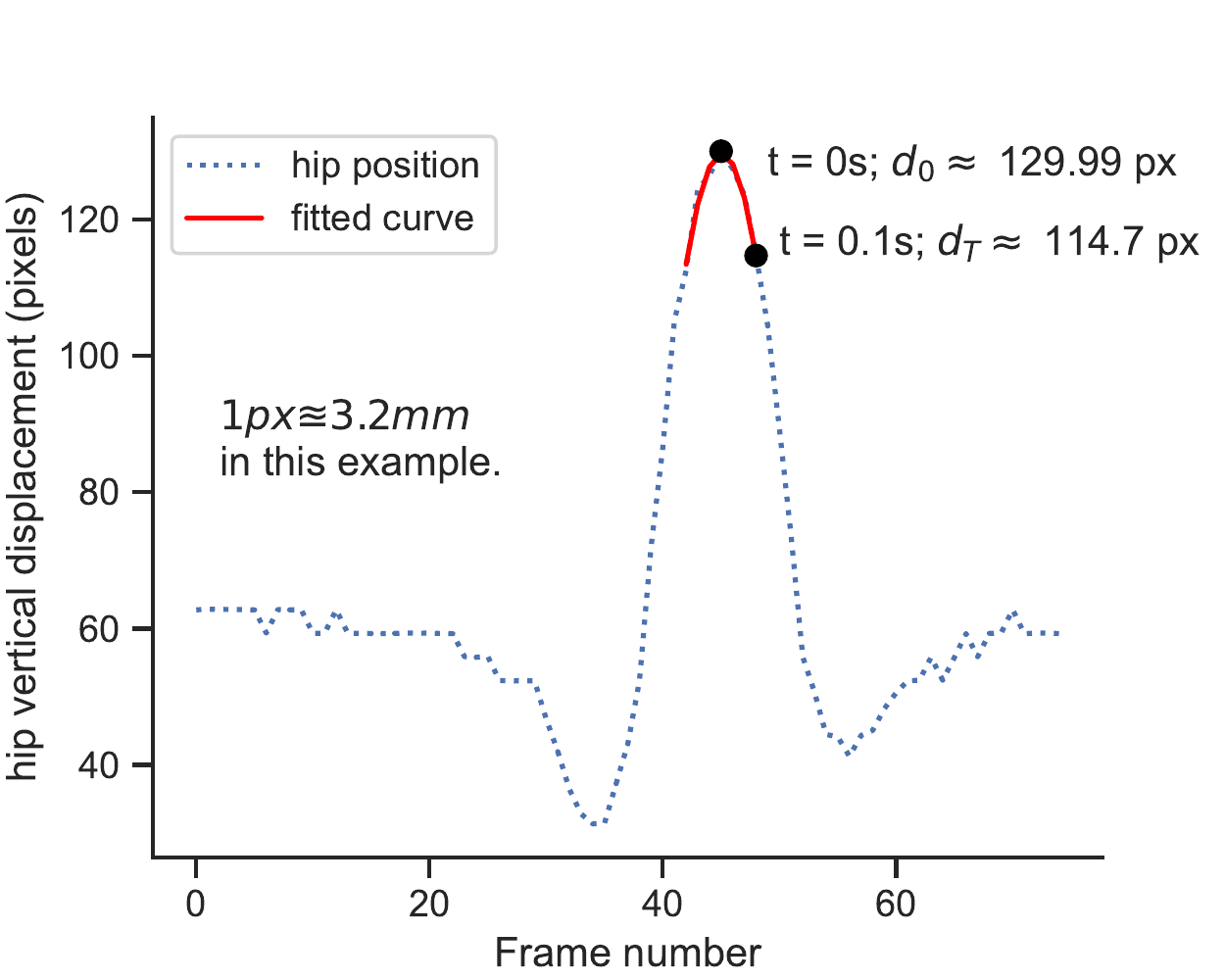}
    \caption{Converting \emph{pixel} to \emph{millimetre} metric scale using gravity as reference.}
    \label{fig:free-fall}
\end{figure}

\textbf{\textit{Pixel-to-Metric (PTM) Conversion}} was performed based on the `free-fall' of the centre of mass during a vertical jump. PTM uses $g$, the universal acceleration due to gravity as reference as proposed in \cite{gravity_ref}. From Newton's law of motion, the motion of a rigid body\footnote{Although the human body is not perfectly rigid, the deformations around the centre of mass are negligible in this instance.} in free fall is described by
\begin{equation}
    d(t) = d_0 + v_0t + \frac{1}{2}gt^2
    \label{eq:gravity}
\end{equation}
where $d_0$ is the initial position in \emph{metres (m)}, $v_0$ is the velocity in \emph{m/s}, and $t$ is the elapsed time in \emph{seconds (secs)}. We set the free-fall duration, $T$, to depend on total hip vertical displacement, such that the hip's non-free-fall motion is not captured. At the peak, $v_0= d_0 =0$ (Fig. \ref{fig:free-fall}). After $T$ secs free fall, $d_{T}=(500T^2g)$\emph{mm}, such that
\begin{equation}
    (500T^2g)mm = |d_0 - d_{T}|px
\end{equation}
Hence, 1 pixel $\equiv \mathcal{R}$ \emph{mm}, where:
\begin{equation}
    \mathcal{R} = \frac{500T^2g}{|d_0 - d_{T}|}
\end{equation}
From this, we obtained $\mathbf{q_{mm}}$ in \textit{mm} scale as
\begin{equation}
    \mathbf{q}_{mm} = \{\mathcal{R}(\mathbf{q_{px_i})} | i=1,..., N\}
    \label{eq:rescale_gravity}
\end{equation}

\subsection{Quantifying Jump Height}
\label{sec:jump_height}
We measured jump heights directly from the OMC and rescaled MMC time series as the maximum vertical displacement of the fifth metatarsal (small toe). We believe this approach is more straightforward than basing measurements on the flight time of the centre of mass, which may vary based on jump strategy.

%% file: 3-results.tex
\section{Analysis and Results}
\label{sec:results}
The jump height reported for each participant is the mean of all three repetitions performed for each task (Table~\ref{tab:all_jumps}). Each MMC measurement was obtained using the reverse-minmax (RMM) and pixel-to-metric (PTM) approaches as described in Section~\ref{sec:rescaling}. The mean $\mathcal{R}$ across all the participants was 3.43\emph{mm/px}. In cases of errors like the one shown in Fig.~\ref{fig:noisy}, the mean value of $\mathcal{R}$ was used.

\subsection{Analysis}
\label{sec:comparisons}
We consider all jump repetitions from all participants as individual measurements, thereby recording 6 jumps per participant and 96 jumps in total, of which 77 (48 bilateral and 29 unilateral) were valid and used for analysis. For quantitative comparison, we use the intraclass correlation coefficient \cite{shrout1979intraclass} (ICC) and Bland-Altman analysis \cite{bland1986statistical} (BA). ICC and BA are often used for comparing new methods of measurements with a gold standard \cite{my_jump,Webering_2021_CVPR, gavin}.

\begin{figure}
    \centering
    $\begin{array}{ccc}
         \subfloat[]{%
          \includegraphics[width=0.90\linewidth]{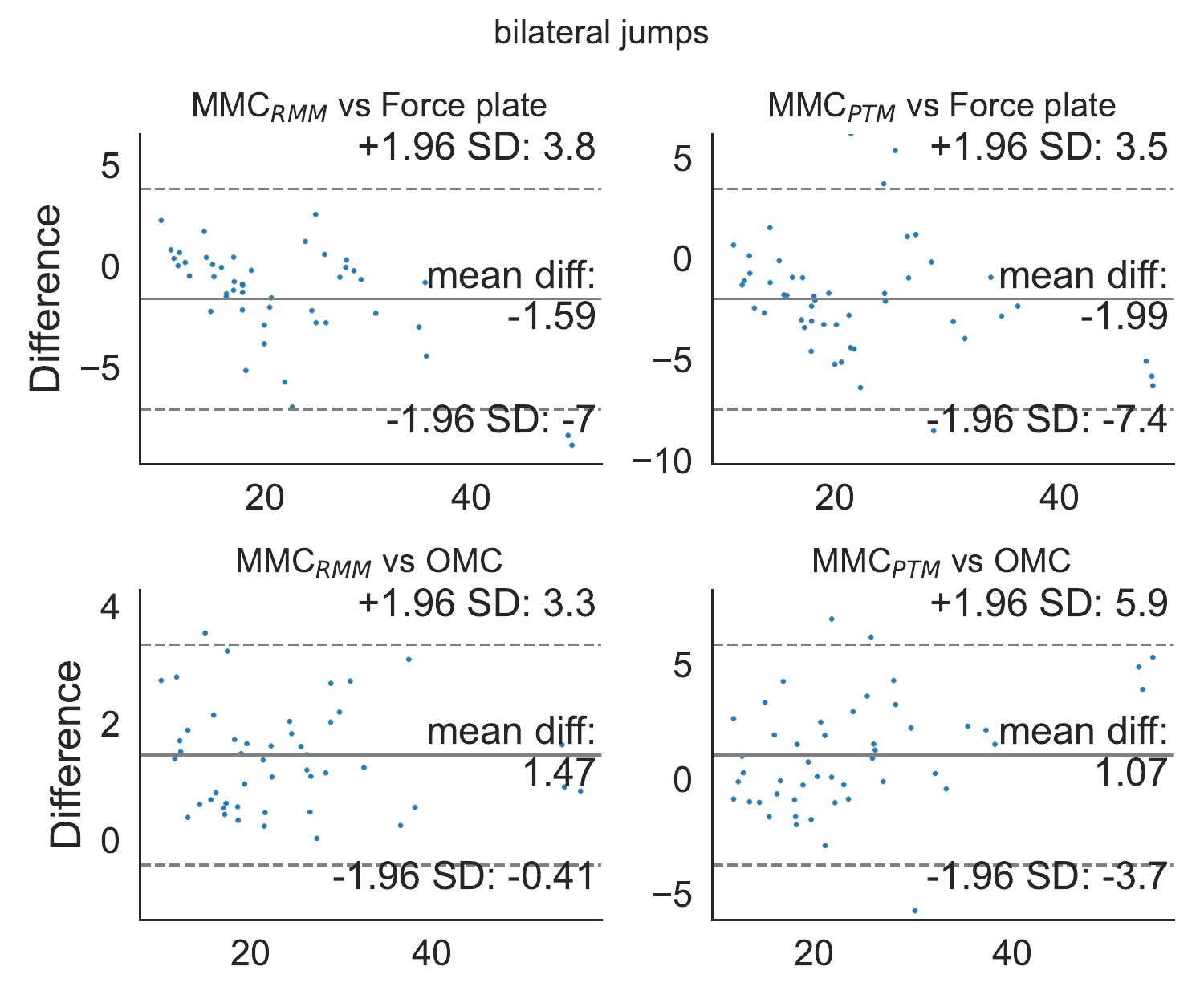}
        } \\
        \subfloat[]{%
          \includegraphics[width=0.90\linewidth]{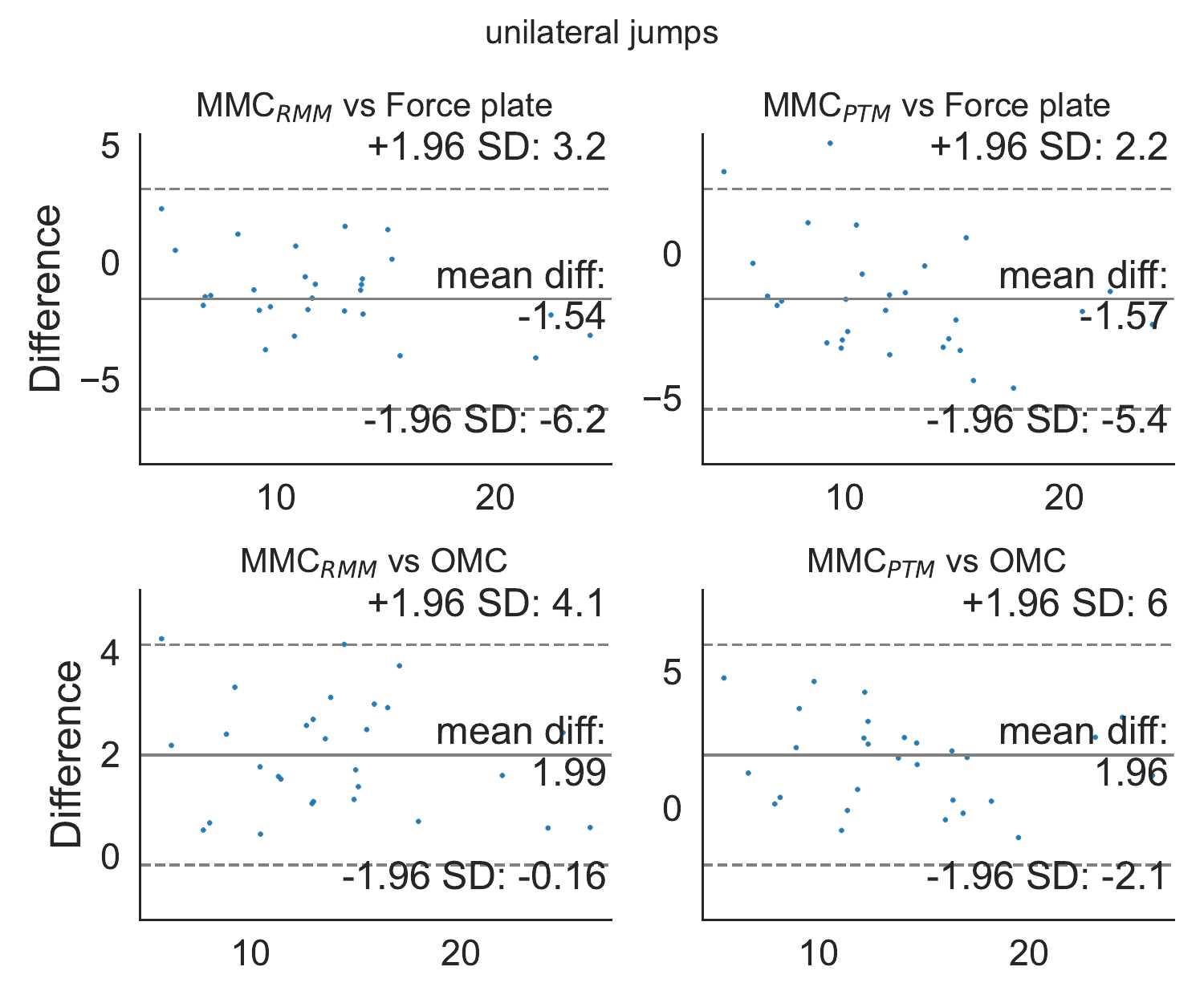}
        } \\
        \subfloat[]{%
          \includegraphics[width=0.90\linewidth]{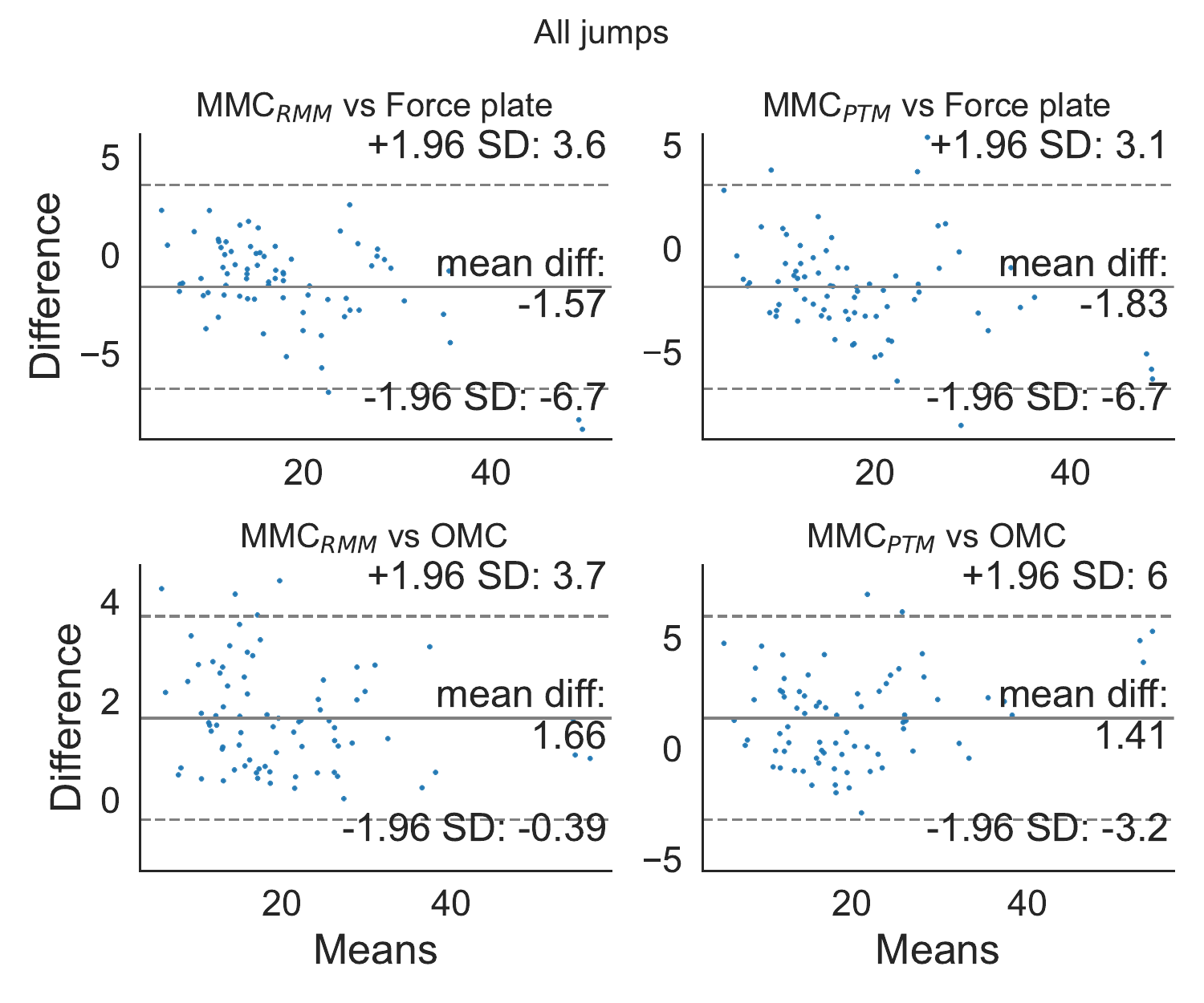}
        }
    \end{array}$
    \caption{Bland-Altman plots for (a) bilateral; (b)  unilateral; (c) all CMJs. Each data point in each scatterplot represents a single jump repetition.}
    \label{fig:baplots}
\end{figure}

\subsubsection{Intra-class Correlation (ICC)}
We took the simultaneous capture of each jump repetition by FP, OMC, PTM and RMM each as a rating. Using the ICC$_{2,1}$, also known as the ``two-way random effects, absolute agreement, single rater/measurement" according to the McGraw and Wong \cite{icc_mcgraw_convention} convention, we compute the inter-rater ICC for four pairs of methods: OMC vs RMM, OMC vs PTM, FP vs RMM, and FP vs PTM, where FP and OMC are taken as ground truths in each case. The ICC $\in [0,1]$ is a measure of the agreement with the ground truths, where a value closer to 1 is preferred.

\begin{table}
\centering
\caption{Intra-session TRR of Force Plate, OMC, and MMC}
\label{tab:trr}
\begin{threeparttable}
    \begin{tabular}{lcc}
    \hline
    \textbf{Technique} & \textbf{Bilateral jumps} & \textbf{Unilateral jumps} \\
    \hline
    Force Plate & 0.98 & 0.97 \\
    OMC & 0.98 & 0.84 \\
    MMC$_{RMM}$ & 0.98 & 0.84 \\
    MMC$_{PTM}$ & 0.95 & 0.89 \\
    \hline
    \end{tabular}
    \begin{tablenotes}
        {\item TRR: test-retest reliability across three repetitions.}
    \end{tablenotes}
\end{threeparttable}
\end{table}

We also computed the intra-rater ICCs to obtain the intra-session test-retest reliability of each measuring technique (shown in Table \ref{tab:trr}) across the three repetitions for each participant. We obtained the ICCs using the Pingouin \cite{vallat_2018} \emph{intraclass\_corr} module.

\begin{table*}[ht]
\centering
\caption{Comparative Analysis and Benchmark}
\label{tab:comparison}
\resizebox{0.99\textwidth}{!}{%
\begin{threeparttable}
    \begin{tabular}{l|llcc|ccc}
    \hline
    & \textbf{Method} & \textbf{Ground Truth} & \textbf{Segmentation} & \textbf{Distance$^1$} & \textbf{ICC$_{2,1}$} & \textbf{bias} (cm) & \textbf{LOA} (cm) \\
    \hline
    \multirow{2}{5em}{SoTA$^2$ (Bilateral)} &
    MMC \cite{Webering_2021_CVPR} & OMC & Auto & Any$^3$ & 0.68 & \textbf{0.15} & [-2.60, 2.90]\\
    & MyJump2 \cite{my_jump} & Force Plate  & Manual & 1 metre & 0.96 & -0.48 & [-2.18, 2.08]\\
    \hline
    \multirow{4}{5em}{Ours (Unilateral)} &
    MMC$_{RMM}$ & OMC & Auto & Any & 0.91 & 1.99 & [-0.16, 4.10] \\
    & MMC$_{PTM}$ & OMC & Auto & Any & 0.86 & 1.96 & [-2.10, 6.00] \\
    & MMC$_{RMM}$ & Force Plate & Auto & Any & 0.84 & -1.54 & [-6.20, 3.20] \\
    & MMC$_{PTM}$ & Force Plate & Auto & Any & 0.88 & -1.57 & [-5.40, 2.20]\\
    \hline
    \multirow{4}{5em}{Ours (Bilateral)} &
    MMC$_{RMM}$ & OMC & Auto & Any & \textbf{0.99} & 1.47 & [-0.41, 3.30] \\
    & MMC$_{PTM}$ & OMC & Auto & Any & 0.97 & 1.07 & [-3.70, 5.90] \\
    & MMC$_{RMM}$ & Force Plate & Auto & Any & 0.95 & -1.59 & [-7.00, 3.80] \\
    & MMC$_{PTM}$ & Force Plate & Auto & Any & 0.93 & -1.99 & [-7.40, 3.50] \\
    \hline
    \multirow{4}{5em}{Ours (BL and UL)} &
    MMC$_{RMM}$ & OMC & Auto & Any & 0.98 & 1.66 & [-0.39, 3.70] \\
    & MMC$_{PTM}$ & OMC & Auto & Any & 0.96 & -1.41 & [-3.20, 6.00] \\
    & MMC$_{RMM}$ & Force Plate & Auto & Any & 0.95 & -1.57 & [-6.70, 3.60] \\
    & MMC$_{PTM}$ & Force Plate & Auto & Any & 0.95 & -1.83 & [-6.70, 3.10]\\
    \hline
    \end{tabular}
    \begin{tablenotes}
        \item $^1$Distance of the participant to the camera. $^2$State of the art as reported in the respective works. $^3$Any distance, as long as the participant is fully visible in the camera view. BL: bilateral; UL: unilateral; RMM: reverse minmax; PTM: pixel-to-metric. Best value for each metric is shown in \textbf{bold} font face.
    \end{tablenotes}
\end{threeparttable}
}
\end{table*}

\subsubsection{Bland-Altman Plots}
The Bland-Altman plots are often used in clinical settings to visualise the agreement between two different methods of quantifying measurements based on bias and limits of agreement (LOA) \cite{giavarina_2015}. The bias $b$ for each MMC measurement technique compared to ground truth is given by the mean of the differences between individual measurements. The LOA is defined as $[c_0,c_1]$, where $c_0=b-1.96SD$, $c_1=b+1.96SD$, and $SD$ is the standard deviation of the differences between the two measurements. At least 95\% of jumps measured with MMC will deviate from OMC by a value within the range $[c_0,c_1]$, where a narrower LOA means better agreement with ground truth. We perform Bland-Altman analysis (Fig.~\ref{fig:baplots}) using statsmodels \cite{seabold2010statsmodels}. 

\subsection{Results}
In this section, we analyse the level of agreement of MMC with OMC and we put this work in context with similar approaches based on ICC, bias, LOA, and simplicity of setup (Table~\ref{tab:comparison}).

\subsubsection{MMC vs OMC}
\label{sec:mmc_vs_omc}
First, the accuracy of MMC in quantifying jump height is evaluated with OMC as ground truth. Both MMC and OMC are measured using the vertical displacement of the toe. As shown in Table~\ref{tab:comparison}, both MMC$_{RMM}$ and MMC$_{PTM}$ achieve results comparable with the work of \cite{Webering_2021_CVPR}, which was also evaluated using OMC equipment. It is worth noting that our PTM approach assumes a simpler setup without manual calibration.

\subsubsection{MMC vs Force Plate}
\label{sec:mmc_vs_force_plate}
The jump height measured from the force plates is taken as the main ground truth in this section. As shown in Table \ref{tab:comparison}, MMC$_{RMM}$ and MMC$_{PTM}$ fall short of the results achieved with MyJump2 \cite{my_jump}, especially during unilateral jumps. This is because the MyJump2 app involves the manual selection of start and end frames of jumps, and also requires subjects to be 1\emph{m} away from the camera. In addition, effective usage of MyJump2 may also require a second party holding the camera. On the other hand, our methods are simpler and more convenient, requiring only a tripod stand and one calibrating jump.

%% file: 4-discussion.tex
\section{Discussion}
\label{discussion}
In this study, we have evaluated 2D markerless motion capture with a single smartphone in quantifying vertical jump height during countermovement jumps. Optical motion capture (OMC) was performed using CODA, and markerless motion capture (MMC) was performed using OpenPose with a single smartphone camera. Jump heights obtained from force plate flight times were used as the first ground truth for evaluating jump height, while OMC was used as the second ground truth. We found that MMC can quantify jump heights with ICC between 0.84 and 0.99 without manual segmentation and camera calibration. For all jumps, the greatest agreement is found between OMC and MMC$_{RMM}$ (LOA [-0.39, 3.70] cm) because Reverse MinMax is performed based on OMC. On the other hand, MMC$_{PTM}$ is more prone to errors (LOA [-3.20, 6.00] cm vs OMC, and [-6.70, 3.10] cm vs FP) since noise in the jump time series is further amplified by the pixel-to-metric conversion factor, $\mathcal{R}$.

Although our proposed methods achieve comparable results, the acceptability of LOA will depend on measures similar to the \emph{minimally important difference} \cite{carton_filan_2020} (MID) in each application context. In order to be acceptable, the LOA should be smaller than the MID. For example, the MID in an elite sports context with high accuracy and precision requirements would be considerably smaller than the MID in recreational athletes.

There are some limitations to our approach. For example, the pixel-to-metric conversion requires a calibrating jump, and movements towards or away from the camera during each task change the pixel-to-metric scale.
In general, the main sources of error we identify in MMC are:
\begin{enumerate}
    \item Video quality. The quality of the video and the amount of clutter in the background affect the confidence of detected keypoints during pose estimation.
    \item Video viewpoint. Accurate detection of body parts is affected by video viewpoint. For example, pose estimation sometimes fails when used for unilateral CMJ in the side view (Fig.~\ref{fig:noisy}). Future studies will explore other views for the unilateral CMJ.
    \item Noise in HPE output. The noise level could be influenced by HPE model accuracy, background clutter, and lighting conditions.
    \item Approximations. Preprocessing steps such as smoothing, segmentation, MMC scaling and pixel-to-metric conversion involve approximations, introducing errors. 
\end{enumerate}
The Force Plate and OMC are also prone to errors due to human factors. For example, OMC coordinates drop to zero when participants' hands or clothes occlude markers. Force values are also affected if participants step outside the force plates momentarily. In cases where such errors were discovered during data collection, the participant was asked to repeat the jump.

%% file: 5-conclusion.tex
\section{Conclusion}
\label{sec:conclusion}
The results of the analyses in this study suggest that markerless motion capture with a single smartphone is promising. However, its use case will depend on the domain-specific minimally important differences (MID). For example, for applications with very small MID, monocular MMC could provide enhanced feedback and/or augmentation for body-worn sensors and markers. On the other hand, for applications such as measuring countermovement jump height, MMC frame-by-frame tracking accuracy is not critical for the method used in this study. Hence, as shown in this study, 2D monocular MMC could potentially replace sensors and physical markers for such applications.

This study focuses on two variants of one motor task with sixteen participants. Future studies will focus on improving and generalising the techniques used to cover a comprehensive range of motor tasks. In addition, the videos used in this study were captured in the side view. Future studies will consider other views and their effects on capture techniques.